# Predictive Analytics for Enhancing Travel Time Estimation in Navigation Apps of Apple, Google, and Microsoft

Pouria Amirian
Ordnance Survey of Great Britain,
Products and Innovation
Southampton, SO16 0AS, UK
+44(0) 7775221531
Pouria.Amirian@os.uk

Anahid Basiri
The University of Nottingham
Nottingham Geospatial Institute (NGI)
Nottingham, NG7 2TU, UK
+44(0)1158467850
Anahid.Basiri@nottingham.ac.uk

Jeremy Morley
Ordnance Survey of Great Britain
Products and Innovation,
Southampton, SO16 0AS, UK
+44(0)2380055168
Jeremy.Morley@os.uk

## ABSTRACT
The explosive growth of the location-enabled devices coupled with the increasing use of Internet services has led to an increasing awareness of the importance and usage of geospatial information in many applications. The navigation apps (often called "Maps"), use a variety of available data sources to calculate and predict the travel time as well as several options for routing in public transportation, car or pedestrian modes. This paper evaluates the pedestrian mode of Maps apps in three major smartphone operating systems (Android, iOS and Windows Phone). In the paper, we will show that the Maps apps on iOS, Android and Windows Phone in pedestrian mode, predict travel time without learning from the individual's movement profile. In addition, we will exemplify that those apps suffer from a specific data quality issue which relates to the absence of information about location and type of pedestrian crossings. Finally, we will illustrate learning from movement profile of individuals using various predictive analytics models to improve the accuracy of travel time estimation.

## 1. INTRODUCTION
Nowadays, increasing use of location-enabled devices has led to an increasing awareness of the importance and usage of geospatial information in many applications. Google recently reported that over half of trillions of searches on Google.com in recent years happened on smartphones. Nearly one-third of all mobile searches on Google.com are related to location. More importantly, location-related mobile searches are growing 50% faster than mobile searches in general [1].

In the navigation context, use of smartphones and their navigation apps, have been replacing satnav devices for car navigation [3, 20]. In general, navigation apps are more efficient as they can easily access to real-time traffic and incidents information and thus are able to recommend alternative routes based on the current situation of the road network. They are economically more appealing than satnavs and from the user experience, there is no need to buy a single purpose device just for navigation, which needs content or software updates once in a while. In fact, all major smartphone platforms (Android, iOS, and Windows Phone) provide at least one native mobile app for navigation (often called "Maps") that pre-loaded by device manufacturers. The Maps apps, use a variety of available data sources to calculate and predict the travel time as well as several options for routing in public transportation, car or pedestrian modes.

For car and public transportation routing, the apps use available data from multitude of sources like official speed limits of routes, real-time speeds derived from transportation sensors in routes, historical average speed data over certain time periods (sometimes just averages, sometimes at particular times of day), actual travel times from previous users, and real-time traffic and incidents information. In order to make higher accuracy prediction most of the Maps apps, act not only as a consumer of the real-time data but as real-time data generators. In other words, most of the real-time traffic information are generated from smartphone devices by continuously sending information about the location of the devices. A Huge amount of location data from a large number of devices then is used to determine various real-time information about current status of road networks. In addition, the collected data are exploited as historical data for tuning the prediction algorithms for routing and estimation of travel time. That is why, most of the navigation apps, can provide more accurate predictions when connected to the internet. In a nutshell, for car and public transportation navigation and routing modes, the Maps apps utilize many data sources and clever predictive analytics in addition to the geospatial data of road networks in order to provide an accurate estimation of travel time for each user.

However, for pedestrian mode, the main (and in most cases the only) source of estimation of travel time is the (pedestrian) road network. Most of the Maps apps tend to use just a formula with a single set of parameters for all users (pedestrians) to calculate the travel time without considering individual users' movement characteristics such as walking speed. Using a formula for all users is not terribly a bad idea. A pedestrian tends to have inertia in her movement characteristics and she usually doesn't change her walking speed frequently. In addition, real-time traffic and incidents in the car and public transportation networks have little impact on individuals' walking speed. Moreover, usually, there is no public source of information for an average speed of pedestrians over certain time periods. The above reasons seem to be rational enough for using just the geospatial data of road network (i.e. length of the road as weight in the graph) for estimation of travel time for pedestrians. In this case, as the user starts walking, the apps provide an estimated travel time based on an average walking speed and length of the route. The travel time is updated at fixed time intervals (usually, a few seconds depending on the app). The estimated travel time mainly is calculated based on the current location of the user (derived from smartphone's GPS) and its distance to the destination (we call this approach "naïve" approach). In other words, no personalization is done in the calculation of the estimation of travel time; that is why for all people at the same location and with the same destination, Maps apps (of same mobile platform)estimate identical values for the travel time. However, the current mobile apps, can learn about the individual's movement characteristics and provide a personal and more accurate estimation of travel time.

While highly intelligent and sophisticated algorithms are being used for different purposes like providing personalized real-time advertisement and recommendation based on browsing history and even for collecting and providing real-time traffic information, no clever algorithm or approach have been used for pedestrian navigation. In pedestrian mode, the Maps apps do not collect data for each user to learn about the movement characteristics of each individual. In other words, Maps apps (in pedestrian mode), on smartphones are not smart enough to utilize the valuable personal source of data for providing more accurate personalized navigation services.

The current naïve approach of predicting travel time results in the less accurate estimation of duration of travel and as a result, it leads to anxiety for pedestrians especially in multi-modal travels. It was reported that users of pedestrian navigation guidance sometimes feel anxiety because of discrepancy between the estimated and actual time of arrival to destination [11]. The naïve approach is also problematic from battery consumption point of view. The current naïve approach of navigation apps needs continuous receiving signals from several GPS satellites. Today's smartphones achieve their long lasting battery life largely because they can aggressively and quickly enter into and exit from sleep states. Use of GPS prevents this clever way of saving battery life. Although this is not a problem for using smartphones in the car, for pedestrians it is a serious problem.

This paper tries to introduce an approach for using movement profile data of the pedestrians and machine learning techniques to improve the estimation of travel time for each individual. Following are the contributions of the paper:

- We evaluate and discuss the accuracy of estimation of travel time in pedestrian mode and related issues in three major native apps for iOS, Android and Windows Phone (section 3).
- We illustrate the use of predictive analytics (supervised machine learning techniques) in improving the accuracy of estimated travel time for pedestrian navigation by learning from individual movement profile data (section 4).

This paper illustrates applying predictive analytics on data which can be easily collected for pedestrians using their smartphones in order to improve the accuracy of estimated travel time and provide personalized services. In this context, surprisingly none of the major routing and navigation apps for pedestrian mode provide such personalized experience for their users.

It is worth mentioning that all the evaluations and predictive analytics are based on our experiments and observations in Oxford, UK. Therefore, the results and conclusions of this research should not be generalized for other geographic areas.

## 2. RELATED WORK

Evaluation of mapping and navigation apps and services is explained and discussed in several research papers. In [22] authors evaluated the Maps apps based on their consistency of the content. In a related work, Samet et al. [21], assessed and compared almost all native mapping apps and major mapping APIs on all major smartphone platforms from user experience and data representation perspectives. Although the evaluation of Maps apps in this paper is focused on a specific functionality of the apps (pedestrian navigation), we will illustrate that a data quality issue in the content has some major impact in the estimation of travel time for pedestrians.

While personal location data from smartphones has been mainly used to improve or construct map contents [8, 25], customized or personalized routing using GPS data of users is subject of several research projects. The idea of Coolest path algorithm was presented in [24]. The Coolest path algorithm enables multi-criteria personalization based on travel distance, travel time, points of interest, and path simplicity. Delling et al. implemented a framework for generating personalized driving directions by automatically analyzing GPS traces [9]. Via examining routes from GPS logs, Letchner et al. [15] found that drivers took the fastest route, as given by a commercial routing engine, only 35% of the time. They presented a set of methods for including driver preferences and time-variant traffic condition estimates in route planning. Chang et al. [6], developed a personalized router for drivers using trajectory mining technique to select routes that were most familiar to the driver. Ziebart et al. [27] modeled the context-dependent utilities and underlying reasons that people take different actions.

Although the above-mentioned works efficiently illustrated personalized route finding in car mode, the same concepts and approaches can be utilized for developing personalized routing engines for pedestrian modes. Personalization using Landmark-based pedestrian navigation has been discussed in [4, 11, 17]. The authors in Going My Way [7] proposed a personalized route planner for landmark-based pedestrian navigation. The system can identify the landmarks automatically from the personal historical GPS log data and can provide navigational instructions based on the landmarks rather than street names. The same functionality is implemented focusing on quality assessment of OpenStreetMap in [5]. Since the Going My Way system learns from data collected from user's navigation history, the system would have no assistance for the user for places the user has never been to. Amirian et al [2] designed and implemented a mobile app and server-side components for providing personalized pedestrian navigation based on landmarks for tourists. In their work, navigational instructions were personalized for each user based on her movement profile. In [26] pedestrian walking speed was modeled based on pedestrian crossing location, pedestrian individual characteristics (gender, age), roadway characteristics (shoulder width, the number of lanes, crossing facilities, signals), and traffic conditions (traffic volume, average travel speed). Rahman et al. [19] applied queuing theory to model pedestrian movement and estimated pedestrian travel time only at roadway links without considering pedestrian crossings.

The work presented in this paper is focused on the predictive analytics for estimation of travel time at the individual level (personalization of travel time estimation) rather than personalization of routing/navigational instructions or modeling walking speed.

# 3. ESTIMATION OF TRAVEL TIME IN PEDESTRIAN MODE

Pedestrian navigation in all native Maps apps of major smartphone platforms uses the naïve approach. After setting the destination (and starting point which usually is the current location of the user) and start the navigation process, the apps, use GPS periodically (every few seconds) to locate the user and calculate the distance between current location and the destination. Based on these calculations the apps update estimated travel time. In other words, for all user at the same location, the duration of travel to the same destination is identical in existing Maps apps of the same platform. This means the apps do not consider any differences between users (gender, age, walking speed, and so on). Despite the fact that available data can be used to provide more realistic and personalized travel time, it hasn't been used in any major navigation apps in all smartphone operating systems. By collecting individuals' movement profile data, it is possible to use modern predictive analytics methods to provide personalized pedestrian navigation services. In this research, we will illustrate the use of movement profile data to improve the predicted travel time for pedestrians. Using predictive analytics, also it is possible to use GPS sensor less frequently and therefore improve the battery life of the smartphones.

## 3.1 Accuracy of Prediction of Travel Time

We conducted an experiment to evaluate the three native Maps apps from major smartphone operating systems; Android, iOS, and Windows Phone. 39 people (21 men and 18 women) with an average age of 33.2 years participated in the experiment. They walked 48 different routes with an average length of 2.8 km (min= 0.8 km and max= 4.5 km) and each route was traveled at least 5 times in Oxford, UK. All routes were navigated with all of the three Maps apps. We used the default Maps app in iOS 9.3.1 and 9.3.2, Windows Phone 8.1 and 10 and Android 4.4.2, 4.4.4, 5.1.2 and 6.0.1. Figure 1 shows one of the routes and its estimated travel time in three different apps for the same route.

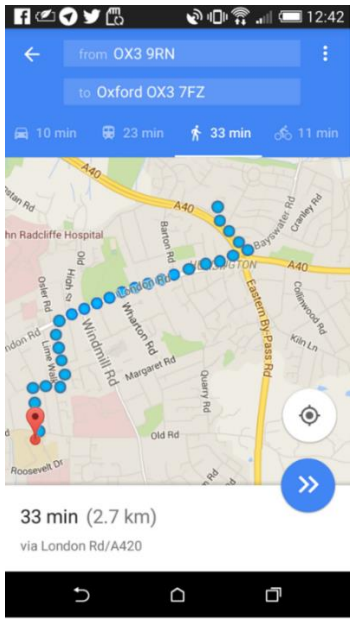

**(a) Android. Estimated travel time 33 minutes.**

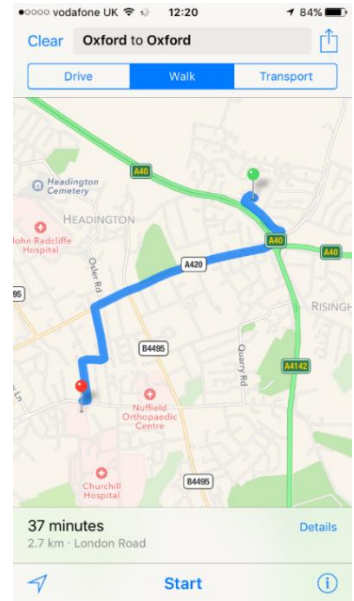

**(b) iOS. Estimated travel time 37 minutes.**

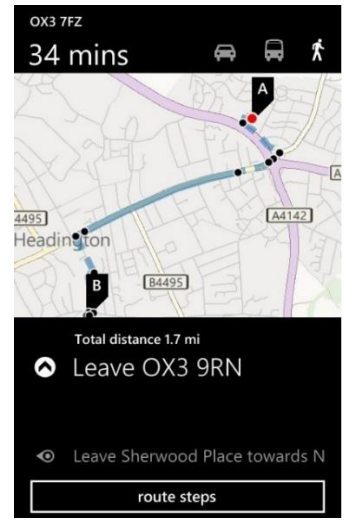

**(c) Windows Phone. Estimated travel time 34 minutes.**

**Figure 1. A route in the experiment in three different Maps apps. Actual travel time for the route based on our experiment is in range of 26-29 minutes.**

In addition to estimated travel time, we asked all participants to measure the actual time of travel using a stopwatch or stopwatch apps. By comparing the actual travel time with predicted travel time of apps we calculated approximation error or relative error for each travel as $e = |t_{actual} - t_{predicted}|/t_{actual}$. Following figures shows the relative error for each app, along with average of relative error and correlation between errors and length of routes.

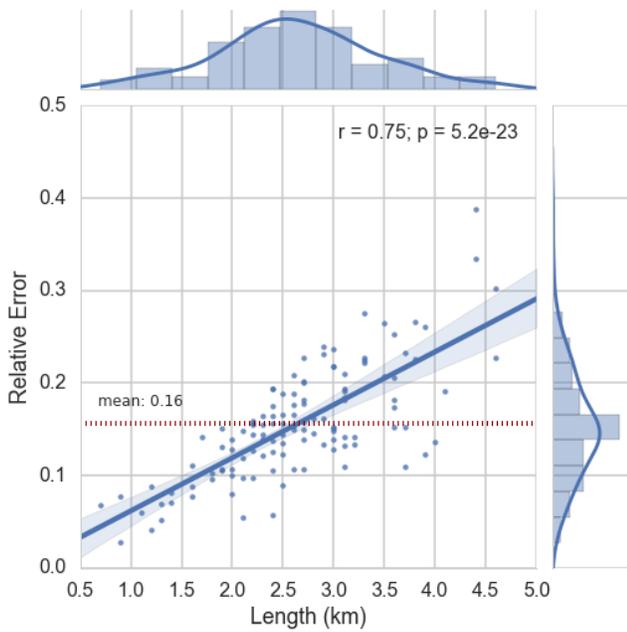

**Figure 2. Relative Error for Android "Maps" app.**

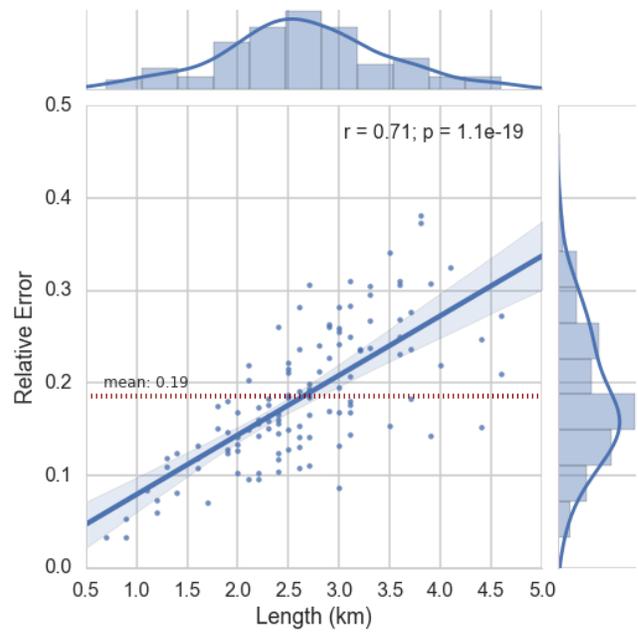

**Figure 4. Relative Error for Windows Phone "Maps" app.**

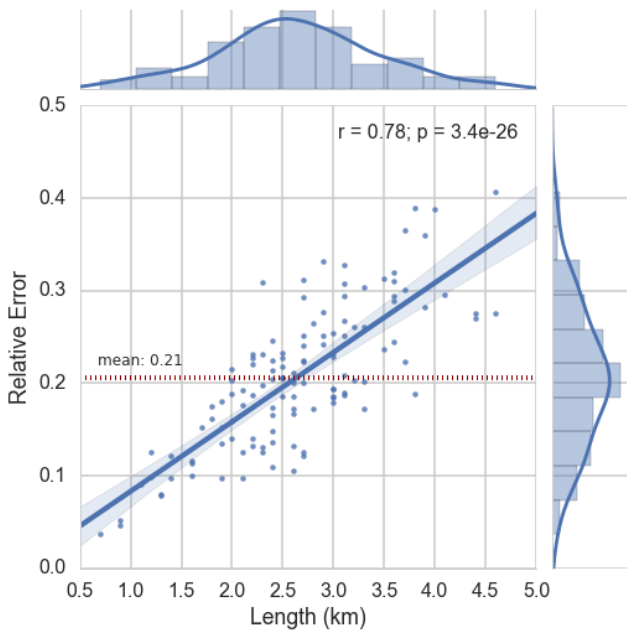

**Figure 3. Relative Error for iOS "Maps" app.**

In all three apps, the length of the route is highly correlated with a relative error in prediction. In other words, as the length of route increases, the relative error linearly grows. In 87% of travels, the "Maps" app on iOS has the largest relative error in estimation of travel time in comparison with Android and Windows Phone apps. Android Maps was better (less relative error) in the estimation of travel time than Windows Phone Maps in 67% of travels.

Estimations mostly were greater than the actual travel time. In just 4 routes (out of 48), the estimation of all apps were less than the actual travel time. In all the mentioned 4 routes, the routes for pedestrian, cross more than 3 highways. In other words, participants in those routes had to wait until the pedestrian lights turned to green. The highway code in the UK includes seven different types of pedestrian crossings each with different types of rules. For example, when cars approach a Zebra crossing, drivers must look out for people waiting to cross and be ready to slow down or stop to let them cross since there are no pedestrian lights for a pedestrians at a zebra crossing.

It seems that a part of the high rate of error in prediction of travel time in pedestrian navigation mode is the lack of information about the type of pedestrian crossing in routing algorithms. In other words, one part of the issue is that the algorithms for routing in pedestrian mode do not differentiate between various types of the pedestrian crossings. With further exploration, we discovered that the algorithms do not use the location of pedestrian crossing in their calculations (seemingly due to lack of information at least in the case study area of this research). With further examination, we found that this is also the case for the online mapping services from Google and Microsoft. As an example, figure 5, shows a route in maps.google.com for crossing a highway to reach to a bus stop (as the initial stage of a multi-modal travel). Based on the suggested route it takes just 2 minutes (146 meters or 479 ft) to reach the bus stop.

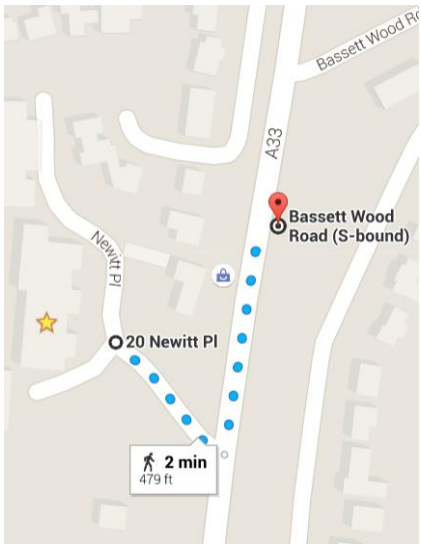

**Figure 5. Suggested route for crossing the A33 road. It is predicted as 2 minutes walking distance.**

As it illustrated in figure 5, it was assumed that pedestrian can cross the highway anywhere. However, since pedestrians must pass the A33 using a pedestrian crossing, they need to walk at least 3 times more than the suggested distance (figure 6).

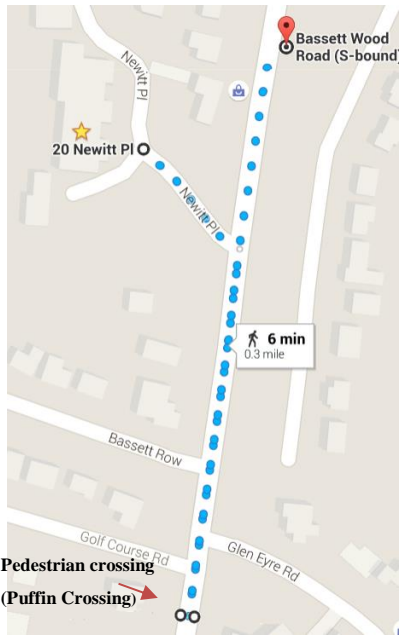

**Figure 6. Actual route for crossing the A33 road. In reality it is 7-8 minutes walking distance.**

In reality, it takes a pedestrian 482 meters to cross the road. Regarding the travel time, if the pedestrian crossing was of type Zebra, that route would take 6 minutes; however, since the pedestrian crossing is of type "Puffin" it takes up to 8 minutes; since the pedestrian has to wait for the traffic light for cars to show red signal. The large difference between estimated and actual travel time can lead to hours of delays in a multi-modal travel. In addition, safe crossing places is a very important characteristic for assessing the walkability of the pedestrian route[14] and an environment's walkability has a major impact of walking speed of individuals [10]. As it illustrated, neither online mapping services nor Maps apps consider walkability of pedestrian routes (at least for case study area of this research).

The location and type of pedestrian crossing are mostly available through OpenStreetMap project. In addition, with the availability of street view data, it is possible to detect the location and type of pedestrian crossing automatically and improve the accuracy of pedestrian navigation. Nevertheless, this issue is related to data quality issue (incompleteness).

Another part of the high rate of error in prediction of travel time in pedestrian navigation services is the lack of personalization and adaptation capabilities. In other words, neither of the native Maps apps, learn from movement profile of the user, weather condition, and personal data in order to improve the accuracy of prediction of travel time for each user. Instead, they just recalculate the distance and travel time between current and the destination locations. Next section is devoted to the approach that we utilized during this research project to overcome the issue.

## 4. PREDICTIVE ANALYTICS FOR IMPROVING THE ACCURACY OF PREDICTION

### 4.1 Dataset

In the mentioned experiment, we collected some other attributes (or features in machine learning terminology) for each travel and participant in addition to the actual time of travel and estimated travel time. Interestingly the dataset of the experiment illustrates the changes of movement behavior based on different attributes. Most of the participants had different travel time for the same routes (and directions) at a different time in same days or in different days. It seems that combinations of environmental and temporal factors cause variable walking speed and thus differentiate between travel time for the same individual for the same route. Apart from environmental and temporal factors, differences between various travel time of various participants are highly correlated with individuals walking characteristics such as average walking speed and the total length of travel during the same day.

The dataset for the predictive analytics contains attributes for temporal and environmental measurements about a certain route for certain person. The dataset composed of four kinds of features:

- User-related such as age, gender, sum of length of journey for current day,
- Temporal and weather information such as time of day, weekday and weather conditions,
- Geospatial attributes of the route such as length and change of elevation within the route,
- Travel time e.g. actual travel time and estimated travel time (estimated by Android's Maps app).

In the dataset of the experiment, all attributes were recorded by each participant except the sum of the length of the journey for current day and change of elevation within the route. For the length of the journey for current day attribute, we utilized (digital) pedometer sensors in smartphones. For each route, the length of the journey of the current day is the number of steps that the user has taken before beginning walking the route. Also for each route we calculated the change of elevation attribute, using the Elevation API of Google Maps. Figure 7, shows the change of elevation for the route in figure 1 and figure 8 illustrates the elevation profile of the same route.

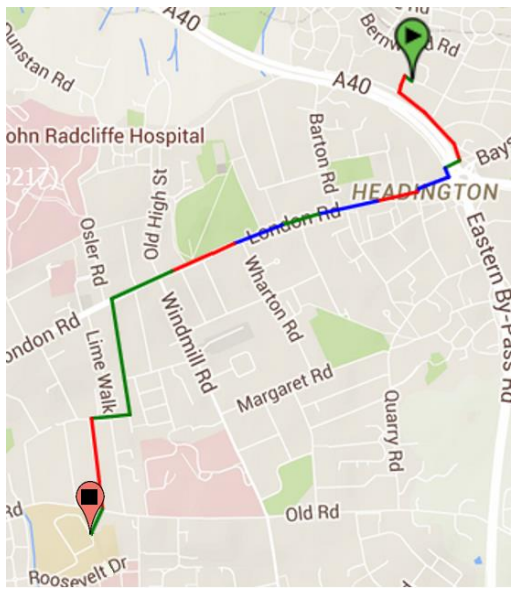

**Figure 7. Change of elevation in a route (the same route which is shown in figure 1). Blue color indicates part of the route with no change of elevation. Green parts specify elevation loss and red is for elevation gain.**

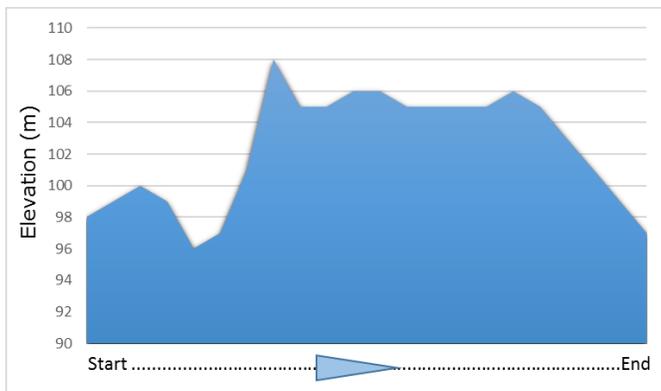

**Figure 8. Change of elevation in the route. Total change in elevation ($|h_{i+1} - h_i|$) is 35 m, elevation loss is 18 m, elevation gain is 17 m.**

Given the sensitivity of detailed movement data of individuals in this research, the dataset contains no raw movement data of individuals. While the anonymity of movement data is hard to achieve [23], it contains patterns that can identify individuals. In fact, De Montjoye et al. [18] studied fifteen months of human mobility data for one and a half million individuals and concluded that human mobility patterns are highly unique. In a dataset where the location is specified every hour and the spatial resolution is coarsely given by antennas, four spatiotemporal points are enough to uniquely identify 95% of the individuals.

## 4.2 Predictive Analytics with All Features

In order to make a personalized prediction for each user, in this research, we trained several predictive models for predicting a correction value based on the environmental, temporal and personal factors. The correction value needs to be added to the estimated value (from Apps) in order to personalize the estimated value of travel time for each user. For predictive modeling, we utilized data from Android devices however same methods can be used with other platforms. From a predictive analytics point of view, we utilize various machine learning algorithms with parameter optimization and cross-validation. Following figure shows the prediction accuracy using $R^2$ (coefficient of determination) for each model.

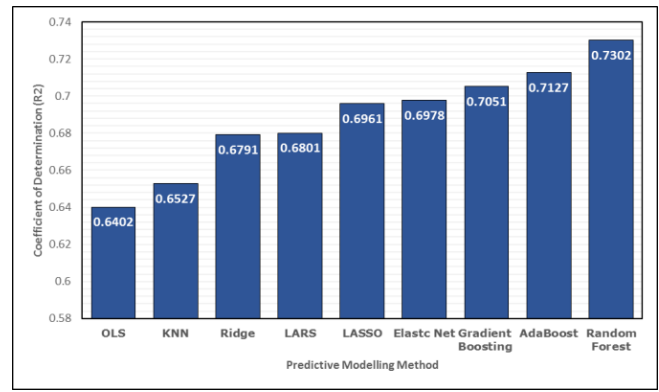

**Figure 9. Prediction accuracy of various machine learning methods for all features. The prediction accuracy is calculated based on $R^2$ (out of sample accuracy) with cross validation.**

As it illustrated in figure 9, even using simple methods like OLS produces a good result. The linear nature of regression problem is the main reason for the relatively high accuracy of the simple method. Since travel time is a linear function of the length of the route, it was expected that the target variable (correction of travel time) could be determined by a linear combination of a subset of potential features. This is illustrated in figure 9 where OLS and penalized linear regression methods produce high accuracy prediction. The penalized linear regression methods, such as Ridge, LASSO, LARS, and Elastic Net generally produce better predictions than OLS solution through a better compromise between bias and variance. As it is shown in figure 9, ensemble learning methods like Gradient Boosting, AdaBoost and Random Forest yield most accurate predictions. Figure 10, shows the relative importance of features in the Random Forest method.

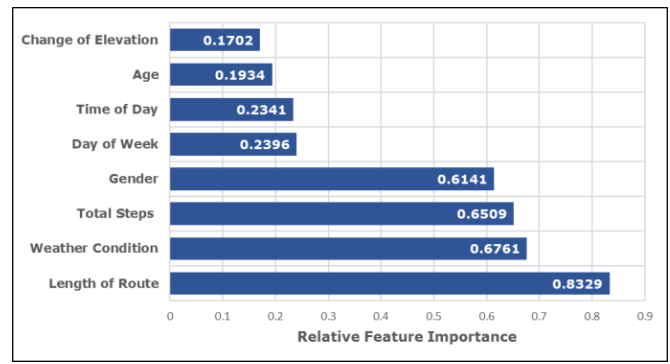

**Figure 10. Relative Feature Importance in Random Forest method (for all features).**

The feature importance is an estimation of prediction strength of each feature [13]. As it expected the length of the route is the most important feature for prediction of the correction value. Weather condition and Total steps (sum of the length of the journey for current day) are two other important features. As it illustrated in figure 10, gender is more important than the age feature. However, this might be due to sample bias in the experiment. In other words, in our experiment the age of participants was between 28 and 38 years (and the average was 33.2) which includes a specific age range and definitely is not representative of the population which uses the smartphones. In the dataset, time of day is highly correlated with total steps and change of elevation has a high value of correlation with the length of the route. This might be the reason for the low importance of the change of elevation and time of day features. With correlated features, strong features can end up with low importance values [12]. This is important especially with regards to the change of elevation since the correlation between length of route and change of elevation in the dataset is 0.7832 which means these two features are collinear.

So far this research showed that with data collection (user profiling) and using predictive modeling techniques it is feasible to provide higher accuracy estimation of travel time for pedestrians. As it shown in figure 9 about 73% of actual correction values can be explained by Random Forest model. Almost all the features in the dataset can be automatically recorded or obtained from the smartphone. However, for myriad reasons accessing to some of the features might not be possible directly. For example, personal information such as age or gender can be obtained in Android SDK using Person class. The Person class has access to age, gender and many other personal information. However, the Person instances have values for age and gender only if the smartphone owner had set up the Gmail or Google+ accounts and filled all necessary information correctly (which is not the case most of the time). So the Person class is not a reliable source for accessing personal data. There are other ways for accessing personal data. Especially with the popularity of social networking apps and unified model of authentication, it is feasible to access the personal information via APIs provided by the major social networking apps (for example using Facebook's Graph API). However, using this approach needs the owner of a smartphone to have an account in the social network and has logged in his/her account. There are some other ways to predict (estimate) the owner's age and gender by using machine learning methods. Recently researchers analyzed the mobile app choices of thousands of Android users to determine the predictability of certain attributes and found that installed apps and app usage pattern can provide highly accurate insight on the user's gender, age, marriage status and even income [16]. However, the methods mentioned in that research is based on the assumption of installing of certain mobile apps.

Nonetheless, the age and gender features are considered highly sensitive and private information and therefore of major concern from privacy, confidentiality, and security point of view. The following section describes the predictive modeling using a subset of features with less concern for privacy.

### 4.3 Predictive Analytics with Less Private Data

As it was mentioned in previous section, gender and to some extent, age are important features for prediction of the correction value. This section illustrates the change in accuracy of predictive modeling using all data excluding age and gender features. The following figure shows the prediction accuracy of the same methods in the previous section.

The results show that the prediction accuracy of all methods is reduced. On average there is 3.5% decrease in prediction accuracy and the prediction accuracy of the Random Forest dropped by 5%.

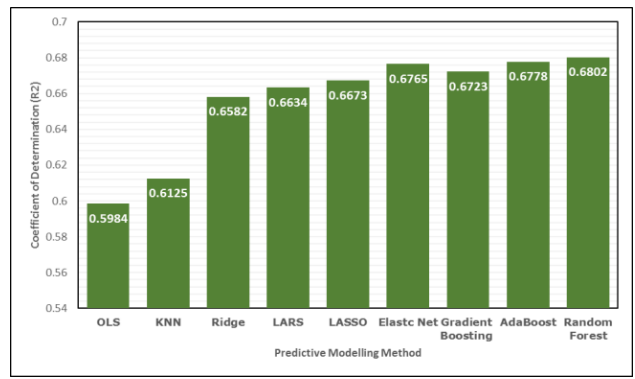

**Figure 11. Prediction accuracy of various machine learning methods (for all features except age and gender) in training of models.**

This indicates the reasonable predictability strength of age and gender (especially gender) on the prediction of the correction value. The following figure shows the feature importance of Random Forest.

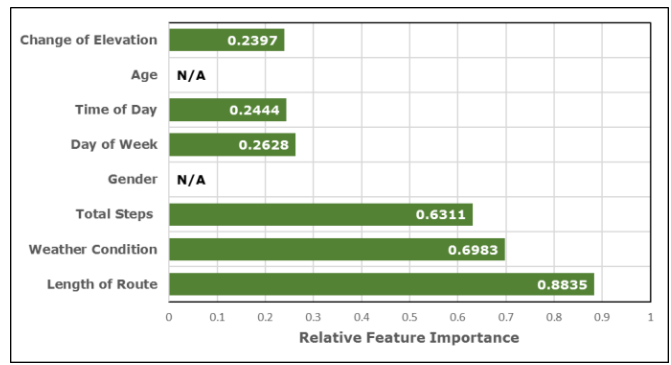

**Figure 12. Relative Feature Importance in Random Forest method (for all but age and gender features).**

The importance order of features is similar to the figure 10. In comparison with figure 10, importance of length of the route, elevation change and weather condition are increased. This means in absence of the private information (gender and age) the predictive modeling method depends more on the other important features in order to keep predicting with high accuracy. This is also the reason for increasing importance value of elevation change. As it mentioned before, the elevation change and length of the route are highly correlated.

### 4.4 Predictive Analytics without Elevation Data

In the dataset, all estimated travel times of both directions of a route was equal for iOS Maps apps. In contrast, in Maps app of Android, and Windows Phone in some cases estimated travel time for both directions of the same route were not identical. The elevation change can cause this phenomenon. In reality, the estimated travel time shouldn't be identical for both directions of the same route. In order to examine the effect of elevation change feature in the prediction of the correction value, we trained the same models on the full dataset excluding age, gender and elevation change features (figure 13).

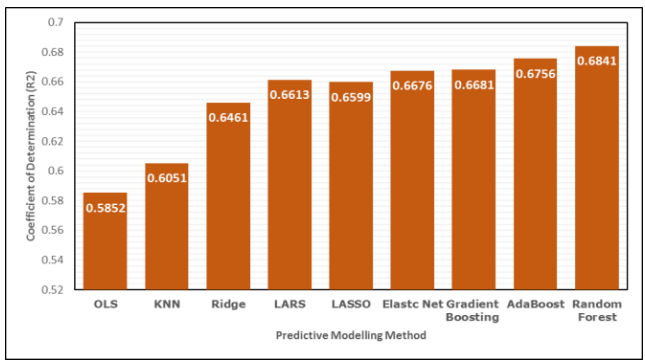

**Figure 13. Prediction accuracy of various machine learning methods (age, gender and elevation excluded).**

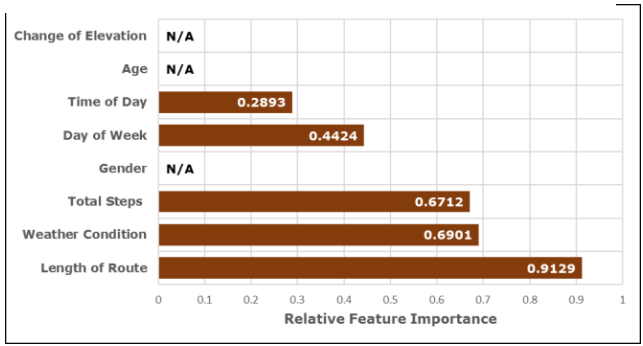

**Figure 14. Relative Feature Importance in Random Forest method (age, gender and elevation excluded).**

The figure 13 is very similar to figure 11 and order to feature importance in figure 14 is the same as figure 12. In comparison with figure 11, there are very small changes in the prediction accuracy of different models. As it illustrated in figure 13, some models even predict with slightly higher accuracy (Random Forest, LARS, and Elastic Net). This might be due to collinearity between elevation change and route length features. In other words, variations in elevation change values can be explained from route length with high accuracy in the dataset. Random noise in the data or confounding features can also cause this interesting result. In addition, we have only used the measurements of Android devices, so the Maps app in Android device might use the slope distance of route instead of horizontal distance. If this is the case the slope distance (and elevation change implicitly) is already included in the model as the length of route feature. Also, this might be the reason for better prediction results of Android Maps app in comparison with Windows Phone or iOS Maps apps. At the other hand, the Android's Maps app might use the horizontal distance but in this specific dataset, including elevation change feature in predictive modeling introduces slightly more variability in the model.

Most accurate routing algorithms must use the slope distance instead of horizontal distance since the slope distance is closer to the route that the pedestrian walks than horizontal distance. However, the dataset of this research is composed of routes with 2.8 km average length in Oxford, UK and Oxford is relatively a flat city. In our dataset, the maximum slope was 9% (in a 16 m segment) in one of the routes when the average slope for the whole route was 3.4%. Average of slopes in a route for all routes was 3.1%. Therefore, there is a negligible difference in estimation of travel time using slope or horizontal distances of routes in our dataset (given the fact that distances in all smartphones are reported using a tenth of kilometers and all travel time are reported in minutes). In other words, whether or not the Android Maps app uses the slope distance needs a larger sample in different geographical regions.

## 5. Conclusion

In this research, we evaluated the pedestrian mode of navigation apps in iOS, Android, and Windows Phone platforms. Through an experiment in Oxford, the UK we explored two major issues of Maps apps for pedestrian navigation.

Lack of information about location and type of pedestrian crossing is the first issue. In this context, we explored that neither of Maps app in iOS, Android and Windows Phone nor commercial online mapping services consider walkability of pedestrian routes in routing algorithms (at least for case study area of this research). This issue is related to data quality and can be solved in various ways; from using crowd-sourcing data like OpenStreetMap to utilizing artificial intelligence for automatic detection of type and location of pedestrian crossing using street view services.

Estimation of travel time for pedestrians is an important aspect of navigation services which in current form can cause several hours of delays in multi-modal travels. Learning from movement profile data of users is an opportunity for providing more accurate services for pedestrians. Unfortunately, neither major Maps apps provide such a personalized service (this is the second issue). In other words, Maps apps on smartphones are not smart enough to learn about the movement profile of their owners to provide higher accuracy services.

As we illustrated in this paper, with predictive analytics it is possible to learn from movement profile of each user and therefore provide more accurate and personalized estimation of travel time in pedestrian mode. With a real-world experiment we showed that using different machine learning algorithms and even without sensitive personal data, estimation of travel time can be significantly improved.